# A Theoretical Framework for Context-Sensitive Temporal Probability Model Construction with Application to Plan Projection

Liem Ngo    Peter Haddawy    James Helwig
Department of Electrical Engineering and Computer Science
University of Wisconsin-Milwaukee
Milwaukee, WI 53201
{liem, haddawy, helwig}@cs.uwm.edu

## Abstract

We define a context-sensitive temporal probability logic for representing classes of discrete-time temporal Bayesian networks. Context constraints allow inference to be focused on only the relevant portions of the probabilistic knowledge. We provide a declarative semantics for our language. We present a Bayesian network construction algorithm whose generated networks give sound and complete answers to queries. We use related concepts in logic programming to justify our approach. We have implemented a Bayesian network construction algorithm for a subset of the theory and demonstrate it's application to the problem of evaluating the effectiveness of treatments for acute cardiac conditions.

## 1 Introduction

Most of our knowledge about the world is highly contextual. The use of context facilitates reasoning by allowing an agent to focus its attention on that portion of its knowledge relevant to a given problem. In automated reasoning the need to use context becomes particularly acute when dealing with probabilistic knowledge. While Bayesian networks provide a relatively efficient method for representing and reasoning with probabilistic information, inference in Bayesian networks remains NP-hard [4]. This complexity becomes particularly problematic as researchers seek to build large models such as those that arise in modeling temporal processes. The most common way to use Bayesian networks for temporal reasoning is to represent time discretely and to create an instance of each time-varying random variable for each point in time.

We can greatly reduce the size of the network models if we can identify some deterministic information and use it as a context to index the probabilistic information. For example, in using Bayesian networks for plan projection, actions are typically represented as nodes in the network [6, ch7], [2, 5]. This often results in networks with large numbers of nodes and large link matrices. The reason is that we need two types of knowledge for each domain variable: a specification of how it is influenced by each action (causal rules), and a specification of how it behaves over time in the absence of actions that influence it (persistence rules). But since when evaluating a plan, the performance of one's own actions is deterministic knowledge — we know whether or not we plan to attempt an action —, actions can be used as context information.

We propose representing a class of Bayesian networks with a knowledge base of probabilistic rules augmented with context constraints. A context constraint is a logical expression that determines the applicability of a probabilistic relation based on some deterministic knowledge. A context-constrained rule has the general form $(P(\text{consequent} \mid \text{antecedents}) = \text{prob}) \leftarrow \text{context}$.

For example, we could represent the effect of a paint action with a set of context-constrained rules, one of which might be $(P(\text{painted}(x,t)) = .99) \leftarrow \text{paint}(x, t-1)$ and the persistence of painted with as set of rules, one of which might be $(P(\text{painted}(x,t) \mid \text{painted}(x, t-1)) = .95)$ Each of these rules has a link matrix half the size of that for a rule representing both the action effect and persistence, and only one of these rules is applicable at any time point.

Although Breese [3] developed a system for reasoning with such context constrained rules, he did not provide a formal semantics for the rules. In previous work [9] we began to address this problem by presenting a formal framework for constructing Bayesian networks from knowledge bases of first-order probability logic sentences. But that work did not allow contextual indexing of probability models and imposed several constraints on the language that limited its expressive power. The objective of this paper is to provide a formal theory for representing context-sensitive temporal probabilistic knowledge and to show how such a representation can be used for efficient probabilistic temporal reasoning.

The rest of this paper is organized as follows. First we provide an overview of the medical problem of evaluat-ing interventions to cardiac arrest, which will be used as a running example throughout the paper. Then we present the representation language and associated semantics for our context-sensitive probabilistic knowledge bases. For a given probabilistic knowledge base, its declarative semantics is defined without reference to any particular reasoning procedure. We provide a procedure that answers probabilistic queries by constructing Bayesian networks from such knowledge bases and we use the formal semantics to prove the procedure both sound and complete. We demonstrate the application of the implemented procedure to our example. Finally we discuss related work.

## 2   The Cardiac Arrest Domain

We illustrate the capabilities of context-sensitive temporal probability model construction by modeling the effects of medications and other interventions on the condition of a patient in cardiac arrest. The goal of treatment is to maintain life and prevent anoxic injury to the brain. Studies show that fewer than 10% of cardiopulmonary resuscitation attempts resulting in survival without brain damage.

The observable variable is the electrocardiogram (EKG) or rhythm strip. During the cardiac arrest, the patient may present with many different cardiac rhythms. While not including all possible rhythms, we consider the range of rhythms most commonly presented: Normal Sinus Rhythm (NSR), Ventricular Fibrillation (VF), Ventricular Tachycardia (VT), Atrial Fibrillation (AF), Super-Ventricular Tachycardia (SVT), Bradycardia (B), Asystole (A).

While patient survival is of primary importance, cerebral damage (CD) must be taken into account. The goal must be to maximize the chance of survival while minimizing the extent of brain damage. The length of time a patient has been without cerebral blood flow (CBF) determines the period of anoxia (POA). If the patient has ineffective circulation for over five minutes, there is a likelihood of sustaining cerebral damage. This damage is persistent and its severity increases as the POA increases.

We consider the two most common medical interventions: cardiopulmonary resuscitation (CPR) and defibrillation (DFIB), the passing of an electric current through the heart, which is used primarily to coordinate contractions in a heart experiencing a dysrhythmia.

Drugs are often used as an initial treatment. They help control the heart rhythm and rate, improve cardiac output and increase blood pressure. Many effective drugs are currently available, of which we choose to model the three most commonly used. Lidocaine (LIDO) is an anti-arrhythmic drug that helps restore a regular rhythm. It is usually used for VT, VF, or to prevent VF. Atropine (ATRO) increases the heart rate during B or A. Epinephrine (EPI) overcomes heart block and helps restore cardiac function.

## 3   The Representation Language

There are two disjoint types of predicates: *context and probabilistic predicates.* Some predicates are timed predicates. A timed predicate always has one attribute indicating the time the associated event or relationship denoted by the predicate occurs. We model only discrete time points and throughout the paper we represent the set of time points by the set of integers. If t is a time point, t+5 denotes the fifth time point after t. If A is a ground timed atom and the time attribute is $t$, we say A (happens) at time $t$.

Context predicates (c-predicates) have value true or false and are deterministic. They are used to describe the context the agent is in and to eliminate unnecessary probabilistic information from consideration by the agent. An atom formed from a context predicate is called a *context atom (c-atom).* A *context literal* is either a c-atom or the negation of a c-atom. A *context base* (CB) is an acyclic normal logic program, which is a set of universally quantified sentences of the form $C_0 \leftarrow L_1, L_2, ..., L_n, n \geq 0$, where $\leftarrow$ stands for implication, comma for logical conjunction, $C_0$ is a c-atom, and the $L_i$ are context literals. We use *completion semantics* proposed by Clark [11] for the semantics of context bases.

Each probabilistic predicate (p-predicate) represents a class of similar random variables. P-predicates appear in probabilistic sentences and are the focus of probabilistic inference processes. An atom formed from a probabilistic predicate is called a *probabilistic atom (p-atom).* Queries and evidence are expressed as p-atoms. In the probability models we consider, each random variable can assume a value from a finite set and in each possible realization of the world, that variable can have one and only one value. We capture this property by requiring that each p-predicate has at least one attribute. The last attribute of a p-predicate represents the *value* of the corresponding variable. For example, the variable *rhythm* of a person can have value *nsr, vf, vt, af, svt, b,* or *a* and can be represented by a two-position predicate, the first position indicating which person and the second indicating that person's type of cardiac rhythm. Associated with each p-predicate $p$ must be a statement of the form $VAL(p) = \{v_1, \ldots, v_n\}$, where $v_1, \ldots, v_n$ are constants. Let $A = p(t_1, \ldots, t_{m-1}, t_m)$ be a p-atom, we use $obj(A)$ to designate the tuple $(p, t_1, \ldots, t_{m-1})$ and $val(A)$ to designate $t_m$. So if A is a ground p-atom then $obj(A)$ represents a concrete random variable or object in the model and $val(A)$ is its value. We also define $Ext(A)$, the *extension of* $A$, to be the set $\{\ p(t_1, \ldots, t_{m-1}, v_i) | 1 \leq i \leq n\}$. If $A$ is an atom of predicate p, then $VAL(A)$ means $VAL(p)$. We assume that p-predicates are typed, so that each attribute of a p-predicate is assigned values in some well-defined domain. We denote the set of all such predicate dec-larations in a knowledge base by *PD*.

Let $A$ be a (probabilistic or context) atom. We define *ground(A)* to be the set of all ground instances of $A$. We assume *mutual exclusivity and exhaustivity*: for each ground p-atom A and in each "possible world", one and only one element in $Ext(A)$ is true. A set of ground p-atoms $\{A_i | 1 \leq i \leq n\}$ is called *coherent* if there does not exist any $A_j$ and $A_{j'}$ such that $j \neq j'$ and $obj(A_j) = obj(A_{j'})$ ( and $val(A_j) \neq val(A_{j'})$ ).

A probabilistic sentence has the form $(P(A_0|A_1,\ldots,A_n) = \alpha) \leftarrow L_1,\ldots,L_m$ where $n \geq 0, m \geq 0, 0 \leq \alpha \leq 1, A_i$ are p-atoms, and $L_j$ are context literals. The sentence can have free variables and each free variable is universally quantified over the entire scope of the sentence. The above sentence is called context-free if $m = 0$. If $S$ is the above probabilistic sentence, we define *context(S)* to be the conjunction $L_1,\ldots,L_m$, *prob(S)* to be the probabilistic statement $P(A_0|A_1,\ldots,A_n) = \alpha$, *ante(S)* to be the conjunction $A_1 \wedge \ldots \wedge A_n$, and *cons(S)* to be $A_0$. Sometimes, we use *ante(S)* as the set of conjuncts and call *cons(S)* the consequent of S. A probabilistic base (*PB*) of a knowledge base is a finite set of probabilistic sentences.

A PB will typically not be a complete specification of a probabiltiy distribution over the random variables represented by the p-atoms. One type of information which may be lacking is the specification of the probability of a variable given combinations of values of two or more variables which influence it. For real-world applications, this type of information can be difficult to obtain. For example, for two diseases $D_1$ and $D_2$ and a symptom $S$ we may know $P(S|D_1)$ and $P(S|D_2)$ but not $P(S|D_1,D_2)$. Combining rules such as noisy-OR and noisy-AND [14] are commonly used to construct such combined influences.

We define *a combining rule* as any algorithm that takes as input a set of *ground* context-free probabilistic sentences with the same consequent $\{P(A_0 \mid A_{i1},\ldots,A_{in_i}) = \alpha_i | 1 \leq i \leq m\}$ such that $\cup_{i=1}^{m}\{A_{i1},\ldots,A_{in_i}\}$ is coherent and produces as output $P(A_0|A_1,\ldots,A_n) = \alpha$, where $A_1,\ldots,$ and $A_n$ are all different and $\{A_1,\ldots,A_n\}$ is a subset of $\cup_{i=1}^{m}\{A_{i1},\ldots,A_{in_i}\}$. We assume that for each p-predicate $p$, there exists a corresponding combining rule in CR, the *combining rules* component of a KB.

A knowledge base (KB) consists of predicate descriptions, a probabilistic base, a context base, and a set of combining rules. We write $KB = \langle PD, PB, CB, CR \rangle$. Figure 1 shows a possible knowledge base for representing the cardiac arrest example introduced in the previous section.

We have the following p-predicates: *rhythm*, *cbf*, *poa*, and *cd*. The statement $rhythm(X : Person, V)$ says that the first argument is a value in the domain *Person*, which is the set of all persons and $V$ can take one value in the set $\{nsr, vf, vt, af, svt, b, a\}$.

So $rhythm(john, t, nsr)$ means the random variable *cardiac rhythm of John at time t*, indicated in the language by $obj(rhythm(john,t,nsr))$, is *normal sinus rhythm*, indicated in the language by $val(rhythm(john,t,nsr)) = nsr$. Similar interpretations apply to other statements in $PD$.

$PB$ contains the probabilistic sentences. Due to space limitations, we cannot show all the sentences. We have a sentence for each possible combination of previous *rhythm* value and resulting *rhythm* value in each possible context. We also need sentences that completely describe the relationship between *rhythm* and *cbf*, cerebral blood flow, *cbf* and *poa*, period of anoxia, and *poa* and *cd*, cerebral damage. $CB$ defines the relationships among context information. The clause $NO\_INTER(X,t) \leftarrow \neg DFIB(X,t), \neg CPR(X,t)$ allows us to imply that no intervention is being employed at a particular time if no intervention is specified. The negation in the antecedent encodes a non-monotonic deduction by negation as failure. Finally, genealized Noisy-OR is used as the combining rule.

## 4  Declarative Semantics

It can be difficult for a user to guarantee global consistency of a large probabilisitic knowledge base, especially when we allow context-dependent specifications. We define our semantics so that we need consider only that portion of a knowledge base relevant to a given problem. Thus if part of the knowledge base is inconsistent, this will not affect reasoning in all contexts. We define the semantics relative to an inference session, characterized by a set of evidence and a set of context information.

**Definition 1** *A set of context information $C$ is any set of c-atoms. A set of evidence $E$ is simply a set of p-atoms. We always assume that ground($E$) is coherent.*

An inference session will be concerned with determining the posterior probabiltiy of some p-atoms $Q$ given $E$, within context $C$.

### 4.1  Context

We interpret a probabilistic sentence of the form $(P(A_0|A_1,\ldots,A_n) = \alpha) \leftarrow L_1,\ldots,L_m$ as representing the conditional probability sentence $(P(A_0|A_1,\ldots,A_n,L_1,\ldots,L_m) = \alpha)$, where free variables are universally quantified outside the scope of the probability operator. The reason for distinguishing syntactically between probabilistic conditioning atoms and context conditioning atoms is that they are treated differently. Conceptually the context information $C$ is a set of evidence or observations which we elaborate with the context base $CB$. We do this by applying completion semantics to $C \cup CB$ and restricting the logical consequences to the Herbrand base on the c-predicates. When $CB$ is acyclic, completed($C \cup CB$)

$PD = \{rhythm(X : Person, t : time, V), VAL(rhythm) = \{nsr, vf, vt, af, svt, b, a\};$
$cbf(X : Person, t : time, V), VAL(cbf) = \{present, absent\};$
$poa(X : Person, t : time, V), VAL(poa) = \{none, 1min, 2min, 3min, 4min, 5min, sustained\};$
$cd(X : Person, t : time, V), VAL(cd) = \{none, mild, moderate, severe\}\};$

$PB = \{P(rhythm(X, 0, nsr)) = 0.001, P(rhythm(X, 0, vf)) = 0.74, \cdots$
$P(poa(X, 0, none)) = 0.99, P(poa(X, 0, 1min)) = 0.005, \cdots$
$P(cd(X, 0, none)) = 0.99, P(cd(X, 0, mild)) = 0.005, \cdots$
$P(cbf(X, 0, present)) = 0.99, P(cbf(X, 0, absent)) = 0.01$
$P(rhythm(X, t, nsr)|rhythm(X, t-1, nsr)) = .05 \leftarrow NO\_INTER(X, t-1), EPI(X, t-1)$
$P(rhythm(X, t, nsr)|rhythm(X, t-1, vf)) = .01 \leftarrow NO\_INTER(X, t-1), EPI(X, t-1)$
$P(rhythm(X, t, nsr)|rhythm(X, t-1, vt)) = .01 \leftarrow NO\_INTER(X, t-1), EPI(X, t-1)$
$\cdots$
$P(rhythm(X, t, vf)|rhythm(X, t-1, af)) = .35 \leftarrow DFIB(X, t-1), ATRO(X, t-1)$
$\cdots$
$P(rhythm(X, t, a)|rhythm(X, t-1, vf)) = .15 \leftarrow NO\_INTER(X, t-1), NO\_MED(X, t-1)$
$\cdots$
$P(cbf(X, t, present)|rhythm(X, t, nsr)) = 1.0$
$P(cbf(X, t, absent)|rhythm(X, t, nsr)) = .0$
$P(cbf(X, t, present)|rhythm(X, t, vf)) = .0$
$P(cbf(X, t, absent)|rhythm(X, t, vf)) = 1.0$
$\cdots$
$P(poa(X, t, 3min)|cbf(X, t-1, present), poa(X, t, 2min)) = .0$
$P(poa(X, t, 3min)|cbf(X, t-1, absent), poa(X, t, 1min)) = .0$
$P(poa(X, t, 3min)|cbf(X, t-1, absent), poa(X, t, 2min)) = 1.0$
$\cdots$
$P(cd(X, t, mild)|poa(X, t, 3min), cd(X, t-1, mild)) = .0$
$P(cd(X, t, mild)|poa(X, t, sustained), cd(X, t-1, severe)) = .0$
$P(cd(X, t, mild)|poa(X, t, sustained), cd(X, t-1, mild)) = .98$
$P(cd(X, t, moderate)|poa(X, t, sustained), cd(X, t-1, mild)) = .02$
$\cdots\}$

$CB = \{NO\_INTER(X, t) \leftarrow \neg DFIB(X, t), \neg CPR(X, t)$
$NO\_MED(X, t) \leftarrow \neg LIDO(X, t), \neg ATRO(X, t), \neg EPI(X, t)\}$

$CR = \{Generalized - Noisy - OR\}$

Figure 1: Example knowledge base.

entails the truth or falsity of each c-atom in the Herbrand base. We take completed($C \cup CB$) to hold with probability one, so that for every ground c-atom $C_i$ either $P(C_i) = 0$ or $P(C_i) = 1$. Now computing the posterior probability of p-atoms $Q$ given evidence $E$, within context $C$ amounts to computing $P(Q|E, completed(C \cup CB))$. Since the probability of each $C_i$ is zero or one, we can condition the sentences in $PB$ on completed($C \cup CB$) by simply eliminating those for which the probability of the context is zero and eliminating the context from those for which the probability of the context is one. Consider the following two sentences from our example KB:

$P(rhythm(X, t, nsr)|rhythm(X, t-1, nsr)) = .05 \leftarrow$
$NO\_INTER(X, t-1), EPI(X, t-1)$
$P(rhythm(X, t, vf)|rhythm(X, t-1, af)) = .20 \leftarrow$
$NO\_INTER(X, t-1), ATRO(X, t-1)$

If $C = \{NO\_INTER(john,1), EPI(john,1)\}$ then conditioning the sentences on completed($C \cup CB$) produces $P(rhythm(john, 2, nsr)|rhythm(john, 1, nsr)) = .05$

The ability to condition the $PB$ in this way will be used in defining the relevant portion of the $PB$.

### 4.2 The Relevant Knowledge Base

In a particular inference session, only a portion of the KB is relevant. The relevant part of KB is determined by the given context information and the set of evidence. The set of relevant atoms is the set of evidence, the set of atoms whose marginal probability is directly stated, and the set of atoms influenced by these two sets, as indexed by the context information. In constructing the set of relevant p-atoms, we consider only the qualitative dependencies described by probabilistic sentences. If $(P(A_0|A_1, \ldots, A_n) = \alpha) \leftarrow L_1, \ldots, L_m$ is a ground instance of a sentence in PB and $L_1, \ldots, L_m$ can be deduced from $completed(C \cup CB)$, then that sentence confirms the fact that $A_0$ *is directly influenced by* $A_1, \ldots, A_n$. If, in addition, $A_1, \ldots, A_n$ are relevant atoms then it is natural to consider $A_0$ as relevant. Let $completed(C \cup CB)$ be the completed logic program with the associated equality theory [11] constructed from $C \cup CB$, then we have the following definition of the set of relevant p-atoms.

**Definition 2** *Given a set of evidence E, a set of context information C and a KB, the set of relevant p-atoms (RAS) is defined recursively by: (1) $ground(E) \subseteq RAS$; (2) if S is a ground instance of a probability sentence conforming to type constraints such that context(S) is a logical consequence of $completed(C \cup CB)$ and $ante(S) \subseteq RAS$ then $cons(S) \in RAS$; (3) if a p-atom A is in RAS then $Ext(A) \subseteq RAS$; (4) RAS is the smallest set satisfying the above conditions.*

The RAS is constructed in a way similar to Herbrand least models for Horn programs. Context information is used to eliminate the portion of PB which is not related to the current problem.

**Proposition 1** *Given a set of evidence E, a set of context information C and a KB, RAS always exists.*

**Definition 3** *Given a set of evidence E, a set of context information C and a KB, the set of relevant probabilistic sentences (RPB) is defined as the set of all prob(S), where S is a ground probabilistic sentence such that context(S) is a logical consequence of completed($C \cup CB$), cons(S) $\in$ RAS and ante(S) $\subseteq$ RAS.*

The RPB contains the basic relationships between p-atoms in RAS. In the case of multiple influences represented by multiple sentences, we need combining rules to construct the combined probabilistic influence.

**Definition 4** *Given a set of evidence E, a set of context information C and a KB, the combined relevant PB (CRPB) is constructed by applying the corresponding combining rules to each maximally coherent set of sentences in RPB which have the same atom in the consequent.*

Combined RPB's play a similar role to completed logic programs. We assume that each sentence in CRPB describes all random variables which directly influence the random variable in the consequent. Given a set of evidence E, a set of context information C and a KB, we can construct CRPB.

We define a syntactic property of CRPB which is necessary in constructing Bayesian networks from the KB.

**Definition 5** *A CRPB is completely quantified if (1) for all ground atoms $A_0$ in RAS, there exists at least one sentence in CRPB with $A_0$ in the consequent; (2) and, for all ground sentence S: $P(A_0|A_1, ..., A_n) = \alpha$ in CRPB, for all $i = 0, .., n$, if val($A_i$) $= v$ and $v' \in VAL(A_i), v \neq v'$, there exists another ground sentence S' in CRPB such that S' can be constructed from S by replacing val($A_i$) by $v'$ and $\alpha$ by some $\alpha'$.*

If we think of each ground $obj(A)$, where A is some p-atom, as representing a random variable in a Bayesian network model then the above condition implies that we can construct a link matrix for each random variable in the model. This is a generalization of constraint (C1) in [9]. We do not require the existence of link matrix for every random variable, but only for the random variables that are relevant to an inference problem.

### 4.3 Probabilistic Independence Assumption

Beside the probabilitistic quantities given in a PB, we assume some probabilistic independence relationships specified by the structure of probabilistic sentences. Probabilistic independence assumptions are used in all related work [3, 13, 9] as the main device to construct a probability distribution from local conditional probabilities. Unlike Poole [13], who assumes independence on the set of consistent *"assumable"* atoms, we formulate the independence assumption in our framework by using the structure of the sentences in CRPB. We find this approach more natural since the structure of the CRPB tends to reflect the causal structure of the domain and independencies are naturally thought of causally.

**Definition 6** *Given a set of ground context-free probabilistic sentences, let A and B be two p-atoms. We say A is influenced by B if (1) there exists a sentence S, an atom A' in Ext(A) and an atom B' in Ext(B) such that A' = cons(S) and B' $\in$ ante(S) or (2) there exists another p-atom C such that A is influenced by C and C is influenced by B.*

**Assumption 1** *Given a set of evidence E, a set of context information C and a KB, we can construct CRPB. We assume that if $P(A_0|A_1, ..., A_n) = \alpha$ is in CRPB then for all ground p-atoms B which are not in Ext($A_0$) and not influenced by $A_0$, $A_0$ and B are probabilistically independent given $A_1, ..., A_n$.*

This assumption is more intuitive and probably easier to check (for knowledge base builders) than the d-separation assumption in [9]. It is well known that these two ways of stating probabilistic independence assumptions are equivalent for finite Bayesian networks [12].

**Example 1** *Continuing the cardiac arrest example, rhythm(john, 1, vf) is probabilistically independent of rhythm(john, 3, vf) given rhythm(john, 2, nsr).*

### 4.4 Model Theory

We define the semantics by using possible worlds on the Herbrand base. This approach of characterizing the semantics by canonical Herbrand models is widely used in work on logic programming [7, 8]. In our semantics, the context constraints in the context base $CB$ and the set of context information $C$ act to select the appropriate set of the possible worlds over which to evaluate the probabilistic part of the sentences. In this way they index probability distributions.

The RAS contains all relevant atoms for an inference session. We assume that *in such a concrete situation, the belief of an agent can be formulated in terms of possible models on RAS.*

**Definition 7** *Given a set of evidence E, a set of context information C and a KB, a possible model M of the corresponding CRPB is a set of atoms in RAS such that for all A in RAS, Ext(A) $\cap$ M has one and only one element.*

A probability distribution on the possible models is realized by a probability density assignment to each model. Let P be a probability distribution on the possible models, we define $P(A_1, \ldots, A_n)$, where $A_1, \ldots, A_n$ are atoms in RAS, as $\sum \{P(w)|w$ is a possible model containing $A_1, \ldots, A_n\}$. We take a sentence of the form $P(A_0|A_1, \ldots, A_n) = \alpha$ as shorthand for $P(A_0, A_1, \ldots, A_n) = \alpha \times P(A_1, \ldots, A_n)$, so that probabilities conditioned on zero are not problematic. We say P satisfies a sentence $P(A_0|A_1, \ldots, A_n) = \alpha$ if $P(A_0, A_1, \ldots, A_n) = \alpha \times P(A_1, \ldots, A_n)$ and P satisfies CRPB if it satisfies every sentence in CRPB.

**Definition 8** *A probability distribution induced by the set of evidence E, the set of context information C, and KB is a probability distribution on possible models of CRPB satisfying CRPB and the independence assumption implied by CRPB.*

We define the consistency property only on the relevant part of a KB. Since the entire KB contains information about various contexts, testing for consistency of such a KB may be very difficult.

**Definition 9** *A completely quantified CRPB is consistent if (1) there is no atom in RAS which is influenced by itself and (2) for all $P(A_0|A_1, \ldots, A_n) = \alpha$ in CRPB, $\sum \{\alpha_i | P(A'_0|A_1, \ldots, A_n) = \alpha_i \in CRPB$ and $obj(A'_0) = obj(A_0)\} = 1$.*

Condition (1) rules out cycles and condition (2) enforces the usual probability assignment constraint.

A possible model in temporal frameworks corresponds to a *world history* [10]. The set of world histories is infinite. In practice, particularly for plan projection problems, we typically consider only events occurring over finite periods. We assume that for each temporal reasoning problem there exist two integers $\iota, \tau, \iota \leq \tau$ such that things occurring outside $[\iota, \tau]$ are not of our concern. So, any timed atom in E or C is at a time in the interval $[\iota, \tau]$.

**Definition 10** *Given two integers $\iota, \tau (\iota \leq \tau)$, a set of evidence E, a set of context information C and a KB, the $(\iota, \tau)$-bounded RAS, denoted by $(\iota, \tau)$-RAS is the set $\{A|A \in RAS$ and if A is timed then it is timed at t and $\iota \leq t \leq \tau\}$. The $(\iota, \tau)$-RPB and $(\iota, \tau)$-CRPB are confined versions of RPB and CRPB, correspondingly, on $(\iota, \tau)$-RAS.*

**Definition 11** *Given two integers $\iota, \tau (\iota \leq \tau)$, a set of evidence E, a set of context information C and a KB, a possible $(\iota, \tau)$-model M of the corresponding CRPB is a set of atoms in $(\iota, \tau)$-RAS such that for all A in $(\iota, \tau)$-RAS, $Ext(A) \cap M$ has one and only one element.*

**Definition 12** *Given two integers $\iota, \tau (\iota \leq \tau)$. A probability distribution which is $(\iota, \tau)$-bound induced by the set of evidence E, the set of context information C, and KB is a probability distribution on possible $(\iota, \tau)$-models of CRPB satisfying $(\iota, \tau)$-CRPB and the independence assumption implied by $(\iota, \tau)$-CRPB.*

**Theorem 1** *Given two integers $\iota, \tau (\iota \leq \tau)$, a set of evidence E, a set of context information C, and a KB, if the $(\iota, \tau)$-RAS is finite and the $(\iota, \tau)$-CRPB is completely quantified and consistent then there exists one and only one $(\iota, \tau)$-bound induced probability distribution.*

In one inference session, we can pose queries to ask for the posterior probabilities of some random variables.

**Definition 13** *A complete ground query wrt the set of evidence E and the set of context information C is a query of the form $P(Q) = ?$, where the last argument of Q is a variable and it is the only variable in Q. The meaning of such a query is: find the poterior probability distribution of $obj(Q)$. If $VAL(Q) = \{v_1, \ldots, v_n\}$ then the answer to such a query is a vector $\vec{\alpha} = (\alpha_1, \ldots, \alpha_n)$, where $0 \leq \alpha_i \leq 1, \sum_{i=1}^n \alpha_i = 1$ and $\alpha_i$ is the posterior probability of $obj(Q)$ receiving the value $v_i$.*

**Example 2** *We can pose the complete ground query $P(rhythm(john, 3, V)) = ?$ to the example KB to ask for the posterior probability of John's cardiac rhythm at time 3.*

An inference problem in an inference session involving a KB, a set of evidence E, a set of context information C is characterized by a query. We represent an inference problem by a tuple $\langle P(Q) = ?, E, C, KB \rangle$.

**Definition 14** *Assume that $VAL(Q) = \{v_1, \ldots, v_n\}$ and Q is at time $t, \iota \leq t \leq \tau$. We say $P(Q) = (\alpha_1, \ldots, \alpha_n)$ is a logical consequence of $\langle P(Q) = ?, E, C, KB \rangle$ if for all probability distributions $P^*$ which are $(\iota, \tau)$-bound induced by E, C, and KB and $\forall 0 \leq i \leq n : P^*(\{M|Q_i$ and E are true in $M\}) = \alpha_i \times P^*(\{M|E$ are true in $M\})$, where $Q_i$ is Q after replacing $val(Q)$ by $v_i$.*

**Example 3** *Suppose we have the query $Q = rhythm(john, 3, V)$. There are an infinite number of induced probability distributions but in all of them $P(rhythm(john, 3, nsr)) = 0.41$, $P(rhythm(john, 3, vf)) = 0.09$, and $P(rhythm(john, 3, vt)) = 0.04$. (See the first example in Section 6.) So they are logical consequences of CRPB.*

In order to prove that the answers returned by our query answering procedure are correct, we need the following definition.

**Definition 15** *In the framework $\langle P(Q) = ?, E, C, KB \rangle, P(Q) = (\alpha_1, \ldots, \alpha_n)$ is a correct answer to the complete ground query $P(Q) = ?$ if $P(Q) = (\alpha_1, \ldots, \alpha_n)$ is a logical consequence of $\langle E, C, KB \rangle$.*

## 5   Query Answering Procedure

In this section we present an algorithm for answering a complete query, which is a generalized form of complete ground queries with possible variables in any attribute of the atom. Assume that we are given $\langle P(Q) =?, E, C, KB \rangle$, where $P(Q) =?$ is a complete query. We call the query answering procedure Q-procedure. Q-procedure uses a form of SLD to reason on PB and SLDNF to answer context queries on $C \cup CB$ [11]. The SLD-like portion of Q-procedure is more complex than SLD because it needs to collect all relevant sentences before combining rules can be used. For that purpose, Q-procedure needs to maintain a list of all ground probabilistic sentences relevant to the current query atom. Q-procedure also calls a Bayesian Net belief updating procedure. There are several available procedures for that purpose [12].

In checking for the validity of contexts, Q-procedure frequently calls the SLDNF proof procedure which works on $C \cup CB$ and queries provided by Q-procedure. SLDNF is sound and, for some classes of normal logic programs, complete under completed program semantics. SLDNF is an efficient proof procedure for normal logic programs and is implemented as the inference engine for the Prolog language. The termination of Q-procedure depends on the termination of SLDNF.

Q-procedure has the following steps: build the necessary portion of the Bayesian network, each node of which corresponds to an $obj(A)$, where $A$ is a ground p-atom in RAS; update the network using the set of evidence E; and output the updated belief of the query nodes. The main idea of the algorithm is to build a *supporting network* for each random variable corresponding to a ground instance of an evidence atom or the query. Let $A$ be a ground p-atom and consider the set of all ground p-atoms $B$ such that $A$ is influenced by $B$ in CRPB. The **supporting network** for $obj(A)$ is a Bayesian network consisting of $obj(A)$ and the set of all $obj(B)$, with the relevant "influenced by" relationships represented as links or sequences of links.

Constructing the network by building supporting networks is justified by the fact that atoms which do not influence either the evidence or the query are irrelevant. To build the supporting networks for the evidence, we first generate the set of all ground instances of the evidence p-atoms. Then for each ground instance, we build the supporting network using PB, the set of ground instances of the evidence which have not been explored, and the current net.

The supporting networks are constructed via calls BUILD-NET, which receives as input an atom whose supporting network needs to be explored. It updates the NET, which might have been partially built. The return value of the function is a set of substitutions such that for each substitution there exists a supporting network for the ground instance of the atom corresponding to the substitution. BUILD-NET frequently

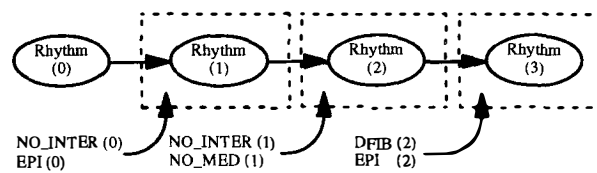

Figure 2: Network for querying heart rhythm.

calls SLDNF to answer queries on CB ∪ C.

**The Correctness of Q-procedure**

In Q-procedure, we only construct the supporting networks, not the entire network for CRPB. We can show that this portion of the network is enough for evaluating the given query.

**Theorem 2** *Given a complete query $P(Q) =?$, where $Q$ is at time $t, \iota \leq t \leq \tau$, a set of evidence E, a set of context information $C$ and a KB. If the COMBINE function always generates finite sets of sentences, CRPB is $(\iota, \tau)$-bound completely quantified and $(\iota, \tau)$-bound consistent and the proof procedure for $C \cup CB$ is sound and complete wrt any query generated by Q-procedure then (1) Q-procedure is sound: the returned answers are correct; (2) Q-procedure is complete: every ground instance of $Q$ which has an answer is returned.*

**The Completeness of Q-procedure**

Our completeness result holds for acyclic rules. The expressiveness of acyclic normal programs is demonstrated in [1]. We hope that acyclic PBs also have an equivalent importance. To the best of our knowledge, PBs with loops are considered problematic and all PB's considered in the literature are acyclic. We need a syntactic criterion called *allowedness* [11] which enables us to prove the soundness and completeness for a large class of KBs.

**Theorem 3** *In any framework $\langle P(Q) =?, E, C, KB \rangle$ where $P(Q) =?$ is a complete query and $Q$ is at a constant time $t, \iota \leq t \leq \tau$, if KB is acyclic (both PB and CB are acyclic), $\langle KB, E, C \rangle$ is allowed, and CRPB is $(\iota, \tau)$-bound completely quantified and $(\iota, \tau)$-bound consistent then Q-procedure is sound and complete.*

## 6   Application

We have implemented a simplified version of Q-procedure called BNG[1]. In this section we demonstrate how the system answers queries for our cardiac arrest domain. Our first example simulates a response to a heart attack. Evidence is presented as the following initial state at time 0: rhythm is ventricular fibrillation, no period of anoxia, no cerebral damage, and cerebral blood flow is present. The actions, administration of Epinephrine at times 0 and 2 and defibrillation at time 2, are represented as context information. Our query is the cardiac rhythm at time 3. Given this inference problem, BNG generates the network shown in Figure 2. The computed posterior probabilities are $P(NSR) = 0.41$, $P(VF) = 0.09$, $P(VT) = 0.04$, $P(AF) = 0.00$, $P(SVT) = 0.01$, $P(B) = 0.00$, $P(A) = 0.44$.

---
[1]The software is available via www at http://www.cs.uwm.edu/faculty/haddawy

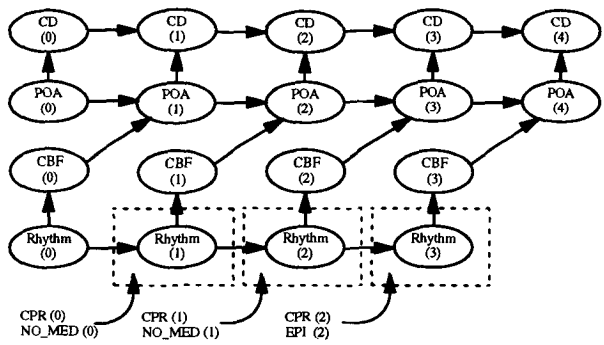

Figure 3: Network for querying cerebral damage.

The next example models a cardiac arrest initiated through drowning. The initial rhythm is asystole, the period of anoxia is known to be 5 minutes, and there is no prior cerebral damage. Treatment consists of Epinephrine administered at time 2 and continued CPR from time 0 to 2. The network generated in response to a query of cerebral damage at time 4 is shown in Figure 3. The computed posterior probabilities of cerebral damage are $P(None) = 0.84$, $P(Mild) = 0.16$.

## 7 Related Work

Our representation language has some similarities to Breese's Alterid [3]. Breese mixes logic program clauses with probabilistic sentences. In contrast, we seperate logic program clauses from probabilistic sentences and use context predicates to select the relevant probabilistic sentences. Breese does not provide a semantics for the knowledge base. As a result, his paper cannot prove the correctness of his query answering procedure. Breese's procedure does both backward and forward chaining. Q-procedure only chains backwards. In that way we can extend the procedure for some infinite domains.

Poole [13] expresses an intention similar to ours: "there has not been a mapping between logical specifications of knowledge and Bayesian network representations ..". He provides such a mapping using probabilistic Horn abduction theory, in which knowledge is represented by Horn clauses and the independence assumption of Bayesian networks is explicitly stated. His work is developed along a different track than ours, however, by concentrating on using the theory for abduction. His theory and ours come to some common points: the acyclicity of sentences, combining rules, completed logic program semantics, and the Bayesian network independence assumption. Our approach concentrates on knowledge base representation and the query answering procedure. We do not allow function symbols in our main results but we think the framework can be extended to include functions.

## Acknowledgements

This work was partially supported by NSF grant IRI-9207262.